\documentclass{article}
\usepackage{ijcai24}

\usepackage{times}
\usepackage{soul}
\usepackage{url}
\usepackage[hidelinks]{hyperref}
\usepackage[utf8]{inputenc}
\usepackage[small]{caption}
\usepackage{graphicx}
\usepackage{amsmath}
\usepackage{amsthm}
\usepackage{booktabs}
\usepackage{algorithm}
\usepackage[switch]{lineno}

\usepackage{lineno,hyperref}
\usepackage{graphicx}
\usepackage{epstopdf}
\usepackage{algorithm}
\usepackage{algorithmicx}
\usepackage{algpseudocode}
\usepackage{amsmath}
\usepackage{amstext}
\usepackage{amssymb}
\usepackage{caption}
\usepackage{subfigure}
\usepackage{color}
\usepackage{multirow}
\usepackage{makecell}

\title{
Scalable Federated Unlearning via Isolated and Coded Sharding
}

\author{
    Anonymous submission
}

\author{
Yijing Lin$^{1,2}$\and
Zhipeng Gao$^{1*}$\and
Hongyang Du$^{4}$\and
Dusit Niyato$^{4}$\and
Gui Gui$^{5}$\and \vspace{0.3em}\\
Shuguang Cui$^{3,2}$\and 
Jinke Ren$^{2,3}$\thanks{Corresponding authors: Jinke Ren and Zhipeng Gao.} 
\affiliations
\vspace{0.5em}
\normalsize{$^1$ State Key Laboratory of Networking and Switching Technology, Beijing University of Posts and Telecommunications\\
$^2$ The Future Network of Intelligence Institute, The Chinese University of Hong Kong (Shenzhen)\\
$^3$ School of Science and Engineering, The Chinese University of Hong Kong (Shenzhen)\\
$^4$ School of Computer Science and Engineering, Nanyang Technological University\\
$^5$ School of Automation, Central South University\\}
\emails \tt{\normalsize{jinkeren@cuhk.edu.cn, gaozhipeng@bupt.edu.cn}}
}

\begin{document}
\maketitle

\begin{abstract}

Federated unlearning has emerged as a promising paradigm to erase the client-level data effect without affecting the performance of collaborative learning models. However, the federated unlearning process often introduces extensive storage overhead and consumes substantial computational resources, thus hindering its implementation in practice. To address this issue, this paper proposes a scalable federated unlearning framework based on isolated sharding and coded computing. We first divide distributed clients into multiple isolated shards across stages to reduce the number of clients being affected. Then, to reduce the storage overhead of the central server, we develop a coded computing mechanism by compressing the model parameters across different shards. In addition, we provide the theoretical analysis of time efficiency and storage effectiveness for the isolated and coded sharding. Finally, extensive experiments on two typical learning tasks, i.e., classification and generation, demonstrate that our proposed framework can achieve better performance than three state-of-the-art frameworks in terms of accuracy, retraining time, storage overhead, and F1 scores for resisting membership inference attacks.

\end{abstract}

\section{Introduction}
\begin{figure}[!t]
  \centering
  \includegraphics[width=0.5\textwidth]{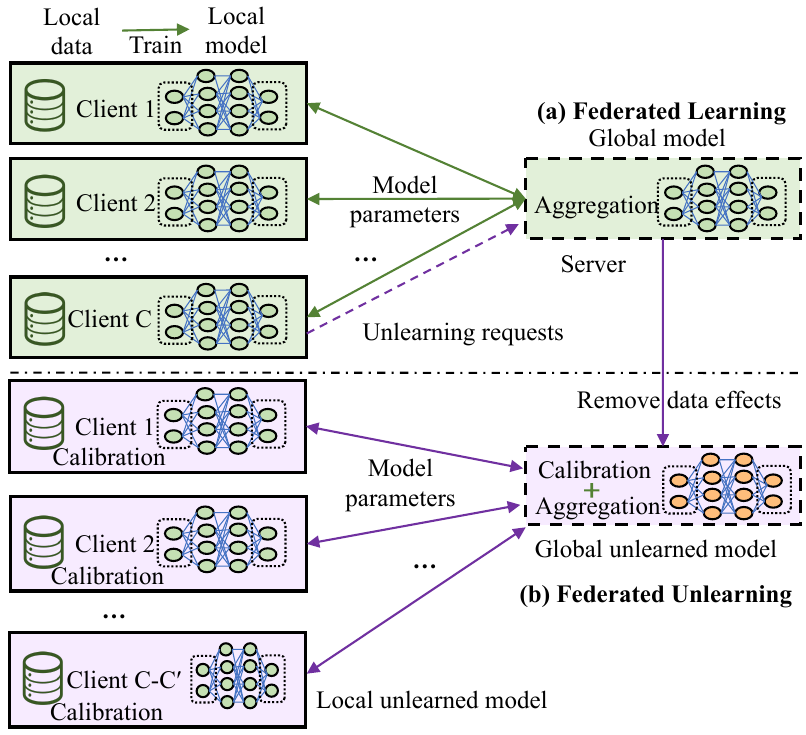}
  \caption{{\bf{Federated Learning vs. Federated Unlearning.}} (a) In federated learning, the clients and the server collaboratively train a global model by exchanging model parameters. (b) In federated unlearning, upon receiving an unlearning request from a specific client $C^{\prime}$, the well-trained global model will be unlearned by using the local models of other clients for calibration, thus removing the corresponding data effect.}  
  \label{fig:comparison}
\end{figure}

With the increasing awareness and stringent regulations of data protection, there is a growing demand for new machine learning technologies that can reap the full benefit of rich data while preserving data privacy.
Federated learning (FL)~\cite{mcmahan2017communication} has become a promising paradigm to collaboratively train a learning model across multiple clients without sharing their local data.~
As shown in Figure~\ref{fig:comparison}(a), each client independently trains a local model using its local dataset and uploads the model parameters to a central server. The server collects the model parameters from clients and broadcasts the aggregated model to all clients. This process is iterated until model convergence.

Despite the great potential of FL, it encounters challenges in data privacy since the shared model parameters still contain some data information. On the other hand, the European Union's General Data Protection Regulation (GDPR) and California Consumer Privacy Act (CCPA) in the United States have released critical requirements for the ``right to be forgotten", which allows clients to erase the data effects from the model parameters trained on their local datasets~\cite{chen2023unlearn,liu2022right,su2023asynchronous}, thus motivates a new computing paradigm called {\it{federated unlearning}}~\cite{liu2021federaser,liu2022right,su2023asynchronous}. As shown in Figure \ref{fig:comparison}(b), each client performs local model calibration and sends the calibrated model parameters to the central server, which removes the data effect from the well-trained FL model via calibration and aggregation. Since the unlearned model does not need to be trained from scratch, the computational overhead can be significantly reduced.


Recent studies on federated unlearning have identified two distinct strategies, including provable and unprovable guarantees. Provable guarantees~\cite{bourtoule2021machine} ensure the complete removal of data effects, while unprovable guarantees indicate that the model has almost forgotten the data effects of specific clients. 
To reduce retraining time, a new framework called FedEraser is proposed in ~\cite{liu2021federaser}, which removes the data effects of specific clients by storing intermediate model parameters on the central server. Although FedEraser achieves provable guarantees and satisfactory unlearning performance, it is short of scalability due to the extensive storage overhead. On the other hand, several orthogonal works~\cite{liu2022right,wu2022federated,wang2022federated} employ various unprovable guarantee-based techniques to enhance the unlearning effectiveness. However, these works mainly optimize loss functions and prune network layers to bound the removal level of data effect, which cannot provide stringent provable guarantees.

To overcome the aforementioned challenges, this paper proposes a scalable federated unlearning framework to efficiently remove the data effects of specific clients while maintaining the model accuracy. Specifically, the entire learning and unlearning process is divided into multiple stages, in which we construct several isolated shards and perform efficient retraining to reduce the storage overhead of the central server. Based on this, we carry out \textit{coded computing} on different shards to further improve the storage efficiency and reduce the training time. To demonstrate the effectiveness of the proposed framework, we conduct extensive experiments on two typical learning tasks, i.e., classification and generation, and compare our proposed framework with three state-of-the-art frameworks, i.e., FedEraser~\cite{liu2021federaser}, RapidRetrain~\cite{liu2022right}, and FedRetrain~\cite{wang2022federated,su2023asynchronous}. Our main contributions can be summarized as follows:

\begin{itemize}
    \item We for the first time introduce a stage-based isolated sharding mechanism to reduce the number of affected clients for federated unlearning. Theoretical analysis proves that the expected time cost for processing unlearning requests can be significantly reduced in both sequential and concurrent cases.
    \item We design a coded computing-based sharding mechanism to improve the system scalability. Specifically, it can improve the storage efficiency by $(1-2\mu)C$ and the throughput by $S/O(C^2 \log^2 C \log \log C)$, where $C$ represents the number of clients, $\mu$ is the proportion of erroneous results, and $S$ is the number of shards.
    \item Experiment results show that the proposed framework can reduce at least 65\% retraining time and 98\% storage overhead while achieving comparable unlearning effectiveness to FedEraser, RapidRetrain, and FedRetrain.
\end{itemize}

\section{Related Work}
To mitigate the data effects of specific clients, machine unlearning is proposed by partitioning training data into isolated slices and performing a joint sharded, isolated, sliced, and aggregated (SISA) method~\cite{bourtoule2021machine,xu2023machine}. Specifically, upon receiving an unlearning request, these slices completely remove the data effects and retrain the learning model
to save computational resources and achieve unlearning with provable guarantees. Besides SISA, some previous works also adopt unprovable guarantee-based methods, such as gradient ascent \cite{chen2023unlearn}, adding unlearning layers via a selective teacher-student formulation \cite{jang2022knowledge}, and transforming the unlearning process into a single-class classification task \cite{yan2022arcane}. Although these methods have achieved good unlearning performance, they need direct access to client data, thus cannot be applied in FL due to data privacy.  

Recent studies have paid attention to removing the data effects of specific clients from well-trained FL models. Specifically, the authors in ~\cite{liu2021federaser} initially propose the concept of federated unlearning and develop a new framework called FedEraser. By using the storage resources of the central server, it can retain intermediate model parameters to achieve unlearning with provable guarantees. To reduce computational overhead while maintaining model accuracy, many unprovable guarantee-based frameworks have also been developed, such as RapidRetrain~\cite{liu2022right}, TF-IDF~\cite{wang2022federated,salton1988term}, FedRecovery~\cite{zhang2023fedrecovery}, KNOT~\cite{su2023asynchronous}, and BFU~\cite{wang2023bfu}. Specifically, RapidRetrain modifies the loss function to accelerate the retraining process and remove the data effects of all clients. In classification tasks, TF-IDF is utilized to unlearn the contributions of specific classes by pruning the most relevant class-discriminative channels~\cite{wang2022federated}. FedRecovery considers differential privacy in the federated unlearning process, where intermediate model parameters from clients are utilized to retrain the global model. KNOT proposes a clustered aggregation-based asynchronous unlearning mechanism to reduce the retraining cost. BFU develops a parameter self-sharing approach to maintain model accuracy while erasing the data effects of clients. While these works have 
achieved impressive unlearning performance via experiments, they mainly consider unprovable guarantees such that the unlearning requirements may not be fulfilled~\cite{bourtoule2021machine}.

\section{Methodology}
In this section, we first describe the federated unlearning framework and then introduce two mechanisms based on isolated sharding and coded computing.
\begin{figure*}[!t]
  \centering
  \includegraphics[width=1.0\textwidth]{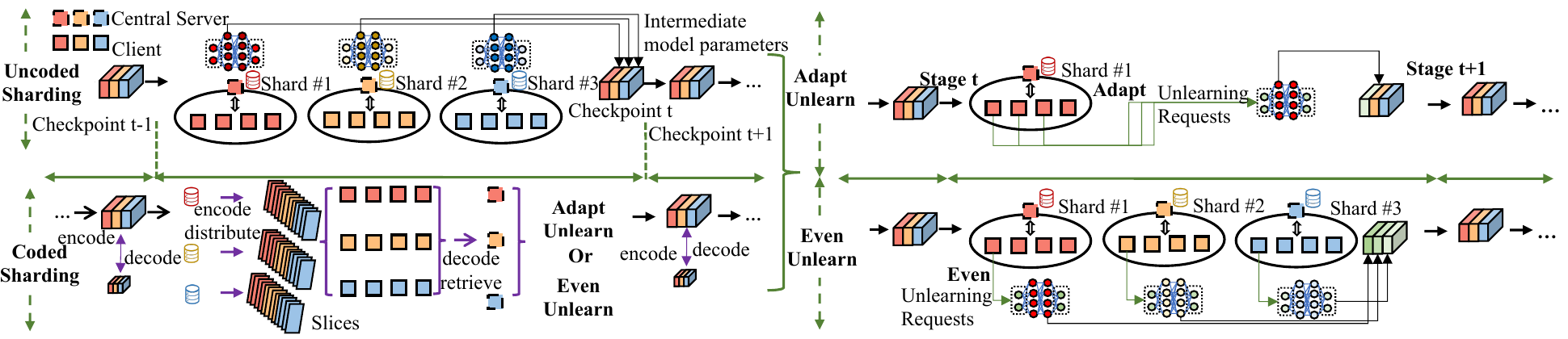}
  \caption{{\bf{Scalable Federated Unlearning Framework}}. For the unlearning requests initiated at different stages, only affected shards perform calibrations to remove the data effects of specific clients. To reduce storage overhead and improve scalability, intermediate model parameters are encoded/decoded as slices for efficient communication between clients and servers.}
  \label{fig:frameowork}
\end{figure*}

\subsection{Scalable Federated Unlearning Framework}
We consider an FL system consisting of $S$ central servers and $C$ distributed clients, denoted by a set $\mathcal{C} = \{C_1,\ldots, C_C\}$. Each client collects a fraction of data and constitutes its local dataset. A shared learning model $\mathbf{w}$ needs to be collaboratively trained across all clients and servers. In the training process, the intermediate model parameters are stored on the servers for subsequent unlearning purposes. To remove the data effects of specific clients, we define a subset of clients, denoted by $\mathcal{C}^{\prime} \subseteq \mathcal{C}$, in which each client needs to be unlearned based on the unlearning requests. Meanwhile, we define $\mathcal{D}^{\prime} \subseteq \mathcal{D}$ as the overall dataset associated with the unlearned clients. Let $\mathbf{w}^{\prime}$ denote the unlearned global model.
Then, according to~\cite{kurmanji2023towards}, the unlearned global model should satisfy
\begin{equation}
    \left\{
    \begin{split}
        &I\left(\mathbf{w}(\mathcal{D}^{\prime});\mathbf{w}^{\prime}(\mathcal{D}^{\prime})\right)=0, \\
        &I\left(\mathbf{w}(\mathcal{D}-\mathcal{D}^{\prime});\mathbf{w}^{\prime}(\mathcal{D}-\mathcal{D}^{\prime})\right)=1,
    \end{split}
    \right.
    \label{eq:requirement}
\end{equation} where $I(\cdot)$ represents the mutual information. 

To achieve (\ref{eq:requirement}), most existing solutions utilize the intermediate model parameters stored on the servers to unlearn the well-trained global model, which inevitably increases the storage overhead of the servers. To solve this problem, we present a new federated unlearning framework, which considers both uncoded and coded sharding to meet adaptive and even unlearning requests. As shown in Figure~\ref{fig:frameowork}, uncoded sharding utilizes a stage-based isolated sharding mechanism to erase the data effect in the global model, which will be illustrated in Section~\ref{sec:stage}. Coded sharding compresses intermediate model parameters into slices, which is able to maintain the storage overhead as the number of clients increases and will be detailed in Section~\ref{sec:coded}.

\subsection{Stage-based Isolated Sharding Mechanism}
\label{sec:stage}
\textbf{Initialization.} Since each client may join or leave the system at any time, we divide the entire learning and unlearning process into multiple stages, where the learning and unlearning operations are performed within each stage.  
Specifically, at each stage, the clients are distributed over multiple shards, denoted by a set $\mathcal{S}$. In each shard, there is a particular server for model aggregation. Therefore, the number of shards is equal to the number of servers, i.e., $S$.
The clients in the shard $s$ are defined by a set $\mathcal{C}_s$. In addition, the dataset and the server associated with the shard $s$ are denoted by $\mathcal{D}_s$ and $J_s$, respectively.

At each stage, the clients in the same shard perform the FedAvg algorithm \cite{mcmahan2017communication} to train a global model. Specifically, each client first trains its local model by $L$ epochs and then transmits the model parameters to the corresponding server for aggregation. Thereafter, the aggregated model parameters are broadcast to the clients for local model update. These steps are iterated for $G$ rounds to obtain a well-trained FL model, which serves as the foundational model for the subsequent unlearning process. Moreover, the server stores the intermediate model parameters of clients in the same shard, which will also be used in the subsequent unlearning process.

\textbf{Preparation}. At each stage, we assume that there are $K$ unlearning requests across the impacted shards, denoted by a set $\mathcal{S}^{\prime}$. In each impacted shard $s_i \in \mathcal{S}^{\prime}$, the clients and unlearning clients are denoted by $\mathcal{C}_{s_i}$ and $\mathcal{C}_{s_i}^{\prime}$, respectively. Let $J_{s_i}$ denote the server associated with the impacted shard $s_i$, which collects intermediate model parameters $\mathbf{w}_{\mathcal{C}_{s_i}}^g, \forall g \in \mathcal{G}$ from the clients in the same shard for storage,
where $\mathcal{G}=\{1,\cdots,G\}$ is the set of the global learning round.
Next, the server $J_{s_i}$ removes the intermediate model parameters associated with the unlearning clients $\mathcal{C}_{s_i}^{\prime}$ from the whole parameter set, as given by
$\mathbf{w}_{s_i}^g=\mathbf{w}_{\mathcal{C}_{s_i}}^g-\mathbf{w}_{\mathcal{C}_{s_i}^{\prime}}^g, \forall g \in \mathcal{G}$. Then, the server aggregates the model parameters of the retained clients to obtain the initial global unlearned model, as given by
\begin{equation}
    \mathbf{w}_{s_i}^{g^{\prime}=0}=\frac{1}{M} \sum_{m=1}^{M} \mathbf{w}_{s_i,m}^g,
    \label{eq:prepare}
\end{equation} 
where $\mathbf{w}_{s_i,m}^g$ is the local model of the retained client $m$ in the shard $s_i$, $M$ is the number of retained clients,  $g^{\prime}$ denotes the unlearning round, and $\mathcal{G}^{\prime}=\{1,\cdots,G\}$ represents the set of all unlearning rounds.
Finally, 
$\mathbf{w}_{s_i}^{g^{\prime}=0}$ is sent to the retained clients in the same shard, i.e., $\mathcal{C}_{s_i}-\mathcal{C}_{s_i}^{\prime}$ 
for subsequent retraining.

\textbf{Retraining.} In the unlearning round $g^{\prime}$, instead of retraining from scratch, each retained client in shard $s_i$ receives the global unlearned model $\mathbf{w}_{s_i}^{g^{\prime}}$ and utilizes its local dataset to update $\mathbf{w}_{s_i}^{g^{\prime}}$ by $\dfrac{L}{r}$ local epochs, where $r$ is a ratio for reducing the number of local retraining rounds~\cite{liu2021federaser}. Then, the updated model parameters are uploaded to the server, which calibrates and updates the global unlearned model by
\begin{equation}
    \mathbf{w}_{s_i}^{{g^{\prime}}+1} = \mathbf{w}_{s_i}^{g^{\prime}} + \frac{1}{M} \sum_{m=1}^M \frac{\Vert \mathbf{w}_{\mathcal{C}_{s_i,m}}^g \Vert}{\Vert \mathbf{w}_{\mathcal{C}_{s_i,m}^{\prime}}^{g^{\prime}}\Vert} \mathbf{w}_{\mathcal{C}_{s_i,m}^{\prime}}^{g^{\prime}}.
    \label{eq:retrain}
\end{equation} 
Here, $g = g^{\prime}$ indicates that $\mathbf{w}_{\mathcal{C}_{s_i,m}^{\prime}}^{g^{\prime}}$ is used to calibrate the local model in the same global learning round. 
Finally, $\mathbf{w}_{s_i}^{g^{\prime}+1}$ is distributed to the retained clients in the shard. This process will be iterated for $G$ rounds.


According to $\eqref{eq:requirement}$,
the provable guarantees of the unlearning process is achieved if the global unlearned model satisfies
\begin{equation}
    \left\{
    \begin{split}
        &I\left(\mathbf{w}_{s_i}^{g}(\mathcal{D}_{s_i}^{\prime});\mathbf{w}_{s_i}^{{g}^{\prime}}(\mathcal{D}_{s_i}^{\prime})\right)=0,\\
        &I\left(\mathbf{w}_{s_i}^{g}(\mathcal{D}_{s_i}-\mathcal{D}_{s_i}^{\prime});\mathbf{w}_{s_i}^{{g}^{\prime}}(\mathcal{D}_{s_i}-\mathcal{D}_{s_i}^{\prime})\right)=1. 
    \end{split}
    \right.
\end{equation} 
To achieve this, it is important to maintain the isolation of the shard in the entire unlearning process, i.e., avoiding cross-shard interactions at each stage. We shall note that there are scenarios where cross-shard interactions are necessary at different stages. In such cases, unprovable guarantee-based methods can be employed. However, it is not the main focus of this paper and will not be discussed herein.

\subsection{Coded Computing-based Sharding Mechanism}
\label{sec:coded}
As stated in \cite{bourtoule2021machine}, an effective federated unlearning framework should be lightweight and scalable, which is able to balance the retraining time, storage overhead, and computational cost for unlearning.
To achieve this goal, we develop a new coded computing-based sharding mechanism, which is composed of two parts, including coded computation and coded reconstruction. 

\textbf{Coded Computation.} In the isolated sharding mechanism, the server stores the uncoded intermediate model parameters $\mathbf{w}_{\mathcal{C}_s}^g, \forall g \in \mathcal{G}$ of the clients in the same shard, which are used to remove the data effects from the global model. Conversely, in the coded computing-based sharding mechanism, the server collects the coded intermediate model parameters, denoted by $ \mathbf{\widetilde{w}}_i, \forall i \in \{1,\cdots,C\}$, which are generated from $\mathbf{w}_{\mathcal{C}_s}^g$ and distributed in clients. Without loss of generality, we utilize the Lagrange interpolation polynomial method~\cite{roth2006introduction} to compute $\mathbf{\widetilde{w}}_i$. Specifically, we first select $S$ different real numbers $\{\omega_1,\ldots,\omega_{S} \}$, where $\omega_s$ is associated with the shard $s$. Then, the Lagrange polynomial with variable $\alpha$ is given by
\begin{equation}
    u(\alpha)=\sum_{s=1}^{S} \mathbf{w}_{\mathcal{C}_s}^g \prod_{j \neq s} \frac{\alpha-\omega_j}{\omega_s-\omega_j}.
\end{equation} 
To distribute the intermediate model parameters over all clients, we further select $C$ different real numbers $\{\alpha_1,\ldots,\alpha_C \}$, where $\alpha_i$ is associated with client $i$ and $C$ is the total number of clients. Based on this, the coded intermediate model parameters stored in client $i$ can be generated at point $\alpha_i$, as
\begin{align}
    \mathbf{\widetilde{w}}_i&=u(\alpha_i)=\sum_{s=1}^{S}\mathbf{w}_{\mathcal{C}_s}^g \prod_{j \neq s} \frac{\alpha_i-\omega_j}{\omega_s-\omega_j}.
    \label{eq:coded_computation}
\end{align}

In the coded computing-based sharding mechanism, each client not only stores a mix of coded intermediate model parameters across different shards but also generates a series of keys. These keys are shared by the clients and the server in each shard, which will be used in the following coded reconstruction. In particular, each client distributes the coded intermediate model parameters to other clients within or outside the shard.

We shall note that coded computation is utilized in the initialization phase described in Section~\ref{sec:stage}. Instead of directly storing the intermediate model parameters, 
the server only collects coded intermediate model parameters from clients for unlearning. Since the size of the coded model parameters is much smaller than that of the uncoded ones, the storage overhead of the server can be significantly reduced.

\textbf{Coded Reconstruction.} In the preparation phase introduced in Section~\ref{sec:stage}, instead of directly utilizing the intermediate model parameters,
the server in each impacted shard utilizes the keys (generated in coded computation) to retrieve and reconstruct these model parameters from clients. This process is akin to decoding a Reed-Solomon code~\cite{roth2006introduction}. Given the evaluations at $C$ distinct points, the server reconstructs the model parameters by decoding a Reed-Solomon code with dimension $S$ and length $C$. The detailed decoding process can be described as follows: {\it{Step 1:}} Let $\{\mathbf{\widetilde{w}}_1,\ldots, \mathbf{\widetilde{w}}_C\}$ denote the corresponding coded slices of the model parameters among clients, which can be obtained according to (\ref{eq:coded_computation}); {\it{Step 2:}} The server uses the keys shared by the clients in the same shard to access these slices; {\it{Step 3:}} The server reconstructs the original model parameters by solving the following linear equation derived from the Reed-Solomon decoding process, as 
\begin{equation}
    \mathbf{\widetilde{w}}_{\mathcal{C}_s}^g = \left[ \begin{array}{cccc}
    1 & \alpha_1 & \cdots & \alpha_1^{S-1} \\
    1 & \alpha_2 & \cdots & \alpha_2^{S-1} \\
    \vdots & \vdots & \ddots & \vdots \\
    1 & \alpha_C & \cdots & \alpha_C^{S-1}
    \end{array} \right]^{-1}
    \cdot
    \left[ \begin{array}{c}
    \mathbf{\widetilde{w}}_1 \\
    \mathbf{\widetilde{w}}_2 \\
    \vdots \\
    \mathbf{\widetilde{w}}_C
    \end{array} \right],
    \label{eq:coded_reconstruct}
\end{equation}
where the matrix formed by $\alpha_i$ is a Vandermonde matrix. It is invertible as long as all elements are distinct. According to \eqref{eq:coded_reconstruct}, the server can accurately reconstruct the intermediate model parameters from the coded slices distributed among all clients. Moreover, since each client only receives a single slice of the coded model parameters, it is difficult for other clients to reconstruct all coded model parameters,
thus preserving the data privacy.

After finishing the decoding process, the server in each impacted shard removes the intermediate model parameters associated with the clients requesting unlearning, initiates the retraining phase, and retrains the global model. The overall federated unlearning process is summarized in Algorithm~\ref{algo:workflow}.

\begin{algorithm}[!t]
  \caption{{Federated Unlearning with Isolated Sharding and Coded Computing}}
  \begin{algorithmic}[1]
  \For{each shard $s \in \mathcal{S}$}
  \State \textbf{Initialization:} // {\tt{Run on clients}}
  \State Perform coded computation via (\ref{eq:coded_computation}).
  \State Distribute $\mathbf{\widetilde{w}}_i$ to the clients within or outside shard.
  \If{$s \in \mathcal{S}$}
  \State \textbf{Preparation:} // {\tt{Run on the server}}
  \State Perform coded reconstruction via (\ref{eq:coded_reconstruct}).
  \State Remove the intermediate model parameters of the unlearning clients.
  \State Obtain the initial unlearned global model via (\ref{eq:prepare}).
  \State \textbf{Retrain:} // {\tt{Run on clients and server}}
  \State Retrain the global model for unlearning via (\ref{eq:retrain}).
  \EndIf
  \EndFor
  \end{algorithmic}  
  \label{algo:workflow}
\end{algorithm}

\section{Theoretical Analysis}
\subsection{Time Efficiency for Isolated Sharding}

\textbf{Sequential Unlearning Requests.} In the sequential setting, each unlearning request is addressed individually. Given $S$ shards and $K$ unlearning requests, the probability that a shard $s$ is selected for retraining $j$ times across $i-1$ unlearning requests is given by
\begin{equation}
    P_s = {i-1 \choose j} \left(\frac{1}{S}\right)^{j} \left(1-\frac{1}{S}\right)^{i-1-j},
\end{equation} where $\dfrac{1}{S}$ is the probability that affects any given shard. Let $\overline{C}_t$ denote the average time cost associated with all shards. Then, the expected time cost for processing all unlearning requests can be estimated by
\begin{align}
    T_{s} &= \sum_{i=1}^{K} \sum_{j=0}^{i-1} \left(\frac{1}{S}\right)^{j} \left(1-\frac{1}{S}\right)^{i-1-j} \overline{C}_t \overset{(a)}{=} K \overline{C}_t,
\end{align} 
where $(a)$ is obtained by the binomial theorem~\cite{bourtoule2021machine}.

\textbf{Concurrent Unlearning Requests.} In the concurrent setting, the unlearning requests are aggregated in a batch for joint processing. Let $\{b_1, \ldots,b_S\}$ denote a set of Bernoulli random variables, where $b_s$ represents whether shard $s$ is affected in the unlearning process. Therefore, $P(b_s=1)=\dfrac{1}{S}$. Accordingly, the expected time cost for processing all unlearning requests can be estimated by
\begin{align}
    T_c &= \mathbb{E} \left(\sum_{s=1}^{S} b_s \overline{C}_t \right)\ \overset{(b)}{=} S \overline{C}_t \left(1-\left(1-\frac{1}{S}\right)^K\right),
\end{align}
where $(b)$ is derived from the expected value of the Bernoulli random variable.

\subsection{Storage Effectiveness for Coded Sharding}
In this work, we use two typical metrics to evaluate the effectiveness of the coded sharding mechanism: 1) \textit{Storage efficiency}, denoted by $\gamma$, which is defined as the ratio of the size of intermediate model parameters to that of the data stored in the server; 2) \textit{Throughput}, denoted by $\lambda$, which characterizes the capacity for processing unlearning requests. The throughput is mainly determined by two factors, i.e., the number of unlearning requests being processed and the associated computational cost. We shall note that the storage efficiency and throughput are the same for the sequential and concurrent settings since only the size of the model parameters is different.

We take the full storage mechanism~\cite{liu2021federaser} as the benchmark, where all the intermediate model parameters are stored in the servers. Therefore, its storage efficiency is set as $\gamma_f=1$. Moreover, since the servers handle all unlearning requests, the throughput is $\lambda_f=1$. For the proposed uncoded sharding mechanism, all clients are divided into $S$ shards. Therefore, its storage efficiency and throughput are given by $\gamma_s=S$ and $\lambda_s=S$, respectively. On the other hand, the proposed coded sharding mechanism is resistant to $\mu C$ erroneous results, where $\mu$ is the proportion of the erroneous results to the coded slices~\cite{roth2006introduction}. Thus, it follows 
\begin{equation}
    2 \mu C \leq C - S,
\end{equation} 
which implies that the upper bound of $S$ is $(1-2 \mu )C$. Accordingly, we have
\begin{equation}
    S \leq \gamma_c \leq (1-2 \mu )C.
\end{equation} Then, given $O(C^2 \log^2 C \log \log C)$ as the additional computational overhead for coded computing~\cite{li2020polyshard}, the throughput of the coded sharding mechanism can be expressed as  
\begin{equation}
    \lambda_c=S/O(C^2 \log^2 C \log \log C).
\end{equation}

\section{Experiments}

\subsection{Experimental Settings}

\textbf{Federated Learning and Unlearning.} We consider a total of 100 clients in our experiments to demonstrate the effectiveness of the proposed framework. In the learning process, only 20 clients are randomly selected in each training round. These clients are divided into 4 shards such that each shard has 5 clients. The numbers of local epochs and training rounds are set as 10 and 30, respectively. In the unlearning process, we consider two types of unlearning requests: 1) \textsf{Even}, where all requests are evenly distributed across shards; 2) \textsf{Adapt}, where all requests are adaptively initiated in one shard~\cite{bourtoule2021machine}.
\begin{figure*}[!t]
  \centering
  \includegraphics[width=1.0\textwidth]{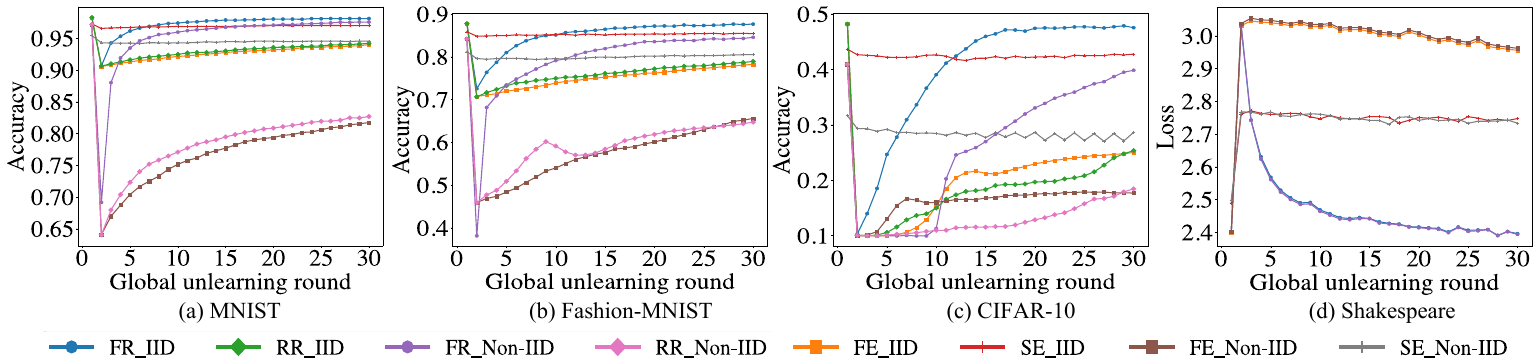}
  \caption{Performance with a single unlearning request.}
  \label{exp:shard_single_request}
\end{figure*}

\begin{table*}[!htbp]
\centering
\begin{tabular}{p{2.8cm}|p{1.35cm}p{1.35cm}|p{1.35cm}p{1.35cm}|p{1.35cm}p{1.35cm}|p{1.35cm}p{1.35cm}}
\hline
\hline
   & \multicolumn{2}{|c}{MNIST}          & \multicolumn{2}{|c}{FMNIST}         & \multicolumn{2}{|c}{CIFAR-10}       & \multicolumn{2}{|c}{Shakespeare}    \\
   \hline
 F1 Score ($\downarrow$)  & IID     & Non-IID     & IID     & Non-IID     & IID     & Non-IID     & IID     & Non-IID     \\
\hline
\textsf{FR} & 0.5455          & 0.5354           & 0.5423          & 0.5434           & 0.5074          & \textbf{0.4742}           & \textbf{0.5426}          & 0.5995           \\
\textsf{FE} & \textbf{0.5450}          & \textbf{0.1953}           & \textbf{0.4659}          & \textbf{0.2803}           & \textbf{0.1649}          & 0.6536           & -               & -                \\
\textsf{RR} & 0.5496          & 0.2289           & 0.5203          & 0.5553           & 0.5264          & 0.6684           & -               & -                \\
\textsf{SE} & 0.5486          & 0.5959           & 0.5447          & 0.5560           & 0.4812          & 0.5141           & 0.6240          & \textbf{0.0392}           \\ 
\hline
Retraining Time ($\downarrow$)  & IID     & Non-IID     & IID     & Non-IID     & IID     & Non-IID     & IID     & Non-IID     \\
   \hline
\textsf{FR} & 565.68          & 563.66           & 556.57          & 558.25           & 573.78          & 571.92           & 2406.96         & 2343.51          \\
\textsf{FE} & 293.84          & 287.40           & 290.90          & 291.94           & 301.57          & 304.22           & 1196.03         & 1213.69          \\
\textsf{RR} & 391.71          & 396.40           & 376.49          & 390.88           & 386.00          & 392.61           & -               & -                \\
\textsf{SE} & \textbf{96.79}           & \textbf{96.51}            & \textbf{96.50}           & \textbf{98.12}            & \textbf{105.05}          & \textbf{103.96}           & \textbf{148.06}          & \textbf{136.95}     \\     

\textbf{Gain} & $\downarrow$ 67.06\% & $\downarrow$ 66.41\% & $\downarrow$ 66.82\% & $\downarrow$ 66.39\% & $\downarrow$ 65.16\% & $\downarrow$ 65.82\% & $\downarrow$ 87.62\% & $\downarrow$ 88.71\% \\

\hline
\hline
\end{tabular}
\caption{F1 score and retraining time in IID and non-IID scenarios.}
\label{tab:extension}
\end{table*}

\textbf{Datasets and Models.} We use four commonly-adopted datasets including MNIST~\cite{lecun1998gradient}, Fashion-MNIST~\cite{xiao2017fashion}, CIFAR-10~\cite{krizhevsky2009learning}, and Tiny Shakespeare~\cite{mcmahan2017communication} for experiments, which are applicable to a diverse range of tasks. On the other hand, we consider two typical learning tasks, i.e., image classification and language generation. To accomplish the two tasks, we use two learning models: 1) Convolutional neural network, which is composed of 2 convolutional, 2 pooling, and 2 fully connected layers. It will be trained on the MINST, Fashion-MINST, and CIFAR-10 datasets, respectively; 2) NanoGPT\footnote{https://github.com/karpathy/nanoGPT}~\cite{radford2019language}, which consists of a 4-layer transformer with 4 attention heads~\cite{vaswani2017attention}, an embedding layer with dimension = 16, a block layer, and a vocabulary with size = 109. In particular, NanoGPT is trained on the Tiny Shakespeare dataset. For data distribution, we consider two data-partition approaches: 1) Independent and identically distributed (IID), where all data samples are randomly divided into 100 equal parts, and each client is assigned with one part; 2) Non-IID. Specifically, for the image classification task, 80\% data samples of each client belong to one primary class, while the remaining data samples belong to other classes~\cite{wang2020optimizing}. For the language generation task, the entire dataset is divided into several unbalanced buckets, and each client is assigned with two buckets to ensure non-IID data distribution across different clients.

\textbf{Baseline Frameworks.} We adopt three state-of-the-art baseline frameworks: 1) FedRetrain (\textsf{FR})~\cite{liu2021federaser}, which retrains the model from scratch to achieve unlearning purposes; 2) FedEraser (\textsf{FE})~\cite{liu2021federaser}, which calibrates model parameters using the historical updates from retained clients; 3) RapidRetrain (\textsf{RR})~\cite{liu2022right}, which employs a diagonal empirical Fisher information matrix to expedite retraining. We name our proposed framework as \textsf{SE}, which is the abbreviation of \textbf{S}harding \textbf{E}raser enabled by isolated and coded sharding.

\textbf{Performance Metrics.} Without loss of generality, the unlearning effectiveness can be evaluated by four metrics, including accuracy, retraining time, storage overhead, and F1 score for resisting membership inference attacks (MIAs)~\cite{shokri2017membership}. Note that F1 score is utilized to verify whether data is successfully unlearned from the model~\cite{bourtoule2021machine,liu2021federaser}. These four metrics align with the fundamental principle outlined in~\cite{bourtoule2021machine}, which focuses on maintaining model accuracy, reducing unlearning time, as well as providing security guarantees. 

\subsection{Experimental Results}

\begin{figure}[!t]
  \centering
  \includegraphics[width=0.5\textwidth]{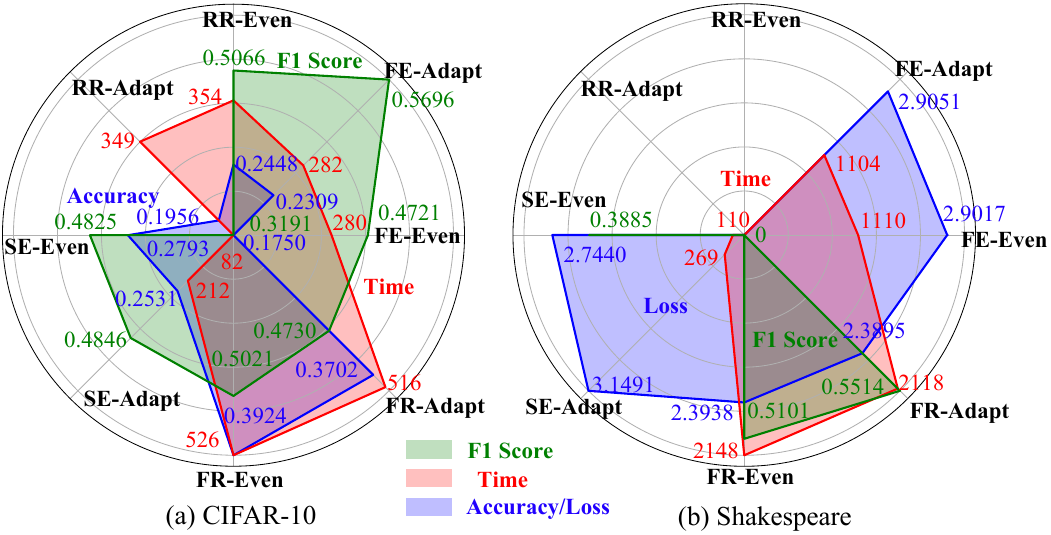}
  \caption{Performance with concurrent unlearning requests.}
  \label{exp:shard_multi_request}
\end{figure}

\begin{figure}[!t]
  \centering
  \includegraphics[width=0.5\textwidth]{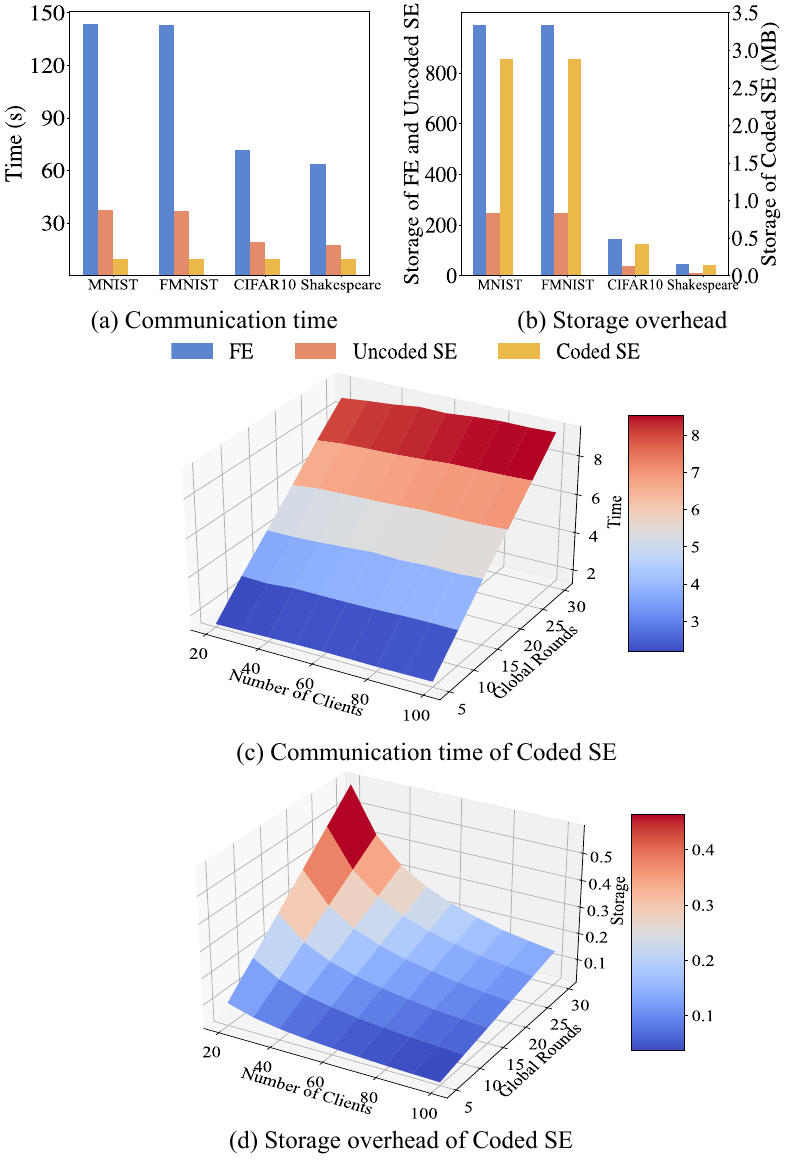}
  \caption{Communication time and storage overhead of different frameworks with concurrent adaptive unlearning requests.}
  \label{exp:store_performance}
\end{figure}

\textbf{Performance with Single Unlearning Request.} We first consider a simple scenario in which each retained client has a single unlearning request. In this scenario, we test the unlearning performance of the four frameworks, and the results are shown in Figure~\ref{exp:shard_single_request}. As can be seen from Figures~\ref{exp:shard_single_request}(a)-(c), the proposed framework \textsf{SE} can achieve comparable accuracy to the baseline framework \textsf{FR} in both IID and non-IID cases. Moreover, \textsf{SE} always outperforms the other two baseline frameworks, i.e., \textsf{FE} and \textsf{RR}. The results on the Tiny Shakespear dataset are very similar, except that the rapid retraining framework \textsf{RR} cannot converge~\cite{su2023asynchronous}. 
On the other hand, since \textsf{FR} retrains the model from scratch and entirely removes data effects of specific clients, its F1 score and accuracy
should be jointly considered for performance comparison. As shown in Table~\ref{tab:extension} and Figure~\ref{exp:shard_single_request}, the F1 score of \textsf{FE} is lower than those of \textsf{FR}, \textsf{RR}, and \textsf{SE}. However, this result is attributed to the reduced accuracy of \textsf{FE}, which cannot be considered as the improved unlearning performance. Instead, our proposed framework \textsf{SE} can achieve comparable F1 scores with \textsf{FR} while maintaining model accuracy. Besides, for the retraining time, our framework \textsf{SE} always surpasses the other three frameworks. For example, in the non-IID case, \textsf{SE} can achieve about 66.41\%, 66.39\%, 65.16\%, and 87.62\% time reduction on the MNIST, Fashion-MNIST, CIFAR-10, and Tiny Shakespeare dataset. Therefore, our proposed framework can significantly reduce the retraining time without sacrificing accuracy.

\textbf{Performance with Concurrent Unlearning Requests.} To demonstrate the scalability of the proposed framework, we further consider a scenario in which each retained client has multiple concurrent unlearning requests. In particular, each request can be evenly or adaptively initiated across different shards. The experimental results are illustrated in Figure~\ref{exp:shard_multi_request}. Specifically, on the CIFAR-10 dataset, the proposed framework \textsf{SE} can outperform \textsf{FE} and \textsf{RR}, and can offer a comparable F1 score to all baseline frameworks. Moreover, it reduces the retraining time by 70\% in the even scenario. Similar trends can be observed in the Tiny Shakespear dataset. These gains are attributed to the fact that \textsf{SE} can significantly reduce the number of affected clients in the unlearning process. As for the adaptive scenario, the variance in data quality across different shards may affect the loss of \textsf{SE} on the Tiny Shakespeare dataset. Nevertheless, its retraining time is still the shortest among all frameworks, demonstrating the effectiveness of the proposed framework in terms of time efficiency.

\textbf{Storage Overhead with Concurrent Unlearning Requests.} Since \textsf{FR} and \textsf{RR} do not utilize storage resources to improve unlearning performance, we only compare the proposed framework \textsf{SE} with the baseline framework \textsf{FE}. We take the concurrent adaptive scenario for experiments. Therein, the communication time for transmitting model parameters includes two parts: 1) Base network delay, which is set as 0.1 second; 2) Model transmission time, which is computed as the ratio of the model size (in bits) and the network data rate (in bit/s). Moreover, to demonstrate the benefit of the coded computing mechanism, we define \textsf{Coded SE} as the framework with isolated and coded sharding, and denote \textsf{Uncoded SE} as the framework with only isolated sharding. In both frameworks, the storage overhead includes the model parameters stored in one shard. 

Due to that the storage overhead is independent of data distribution, we take the IID scenario as an example. Figure~\ref{exp:store_performance}(a) and Figure~\ref{exp:store_performance}(b) show the communication time and storage overhead of  \textsf{FE}, \textsf{Uncoded SE}, and \textsf{Coded SE} with concurrent adaptive unlearning requests. We can observe that \textsf{Coded SE} achieves the smallest storage overhead and the minimum communication time among all frameworks. For example, the storage overhead can be reduced by almost 98\%. On the other hand, we perform experiments on the Shakespeare dataset to investigate the impact of the total number of clients and global rounds on the storage overhead. The results are depicted in Figures~\ref{exp:store_performance}(c)-(d). From these figures, we can see that \textsf{Coded SE} shows a sharp reduction in storage overhead with a slight increase in communication time as the number of clients increases. This result is attributed to the increased division of model parameters into more shards for storage when the number of clients increases. Also, there exists a marginal increase in the distribution and retrieval time when the server utilizes the model parameters for unlearning. 

\section{Conclusion}

In this paper, we have developed a scalable federated unlearning framework by introducing two mechanisms based on isolated sharding and coded computing. The isolated sharding mechanism divides distributed clients into multiple shards and removes the data effects according to sequential or concurrent unlearning requests. The coded computing mechanism extends the scalability of the framework by encoding, decoding, distributing, and retrieving intermediate model parameters. By doing so, the storage overhead of the central server can be significantly reduced. Theoretical analysis and experimental results demonstrate the effectiveness of the proposed framework as compared with three baseline frameworks. Future works may consider integrating unlearning layers into model architectures to achieve cross-shard unlearning.

\bibliographystyle{named}
\bibliography{ijcai24}

\end{document}